\documentclass[11pt,a4paper]{article}
\usepackage[hyperref]{eacl2021}
\usepackage{times}
\usepackage{latexsym}

\usepackage{hyperref}
\usepackage{booktabs}       
\usepackage{amsfonts}       
\usepackage{graphicx}
\usepackage{multirow}
\usepackage{amsmath}
\usepackage{mathtools}
\usepackage{amssymb}
\usepackage[inline]{enumitem}

\usepackage{microtype}

\newcommand{\ra}[1]{\renewcommand{\arraystretch}{#1}}

\aclfinalcopy 

\title{Scalable Evaluation and Improvement of Document Set Expansion \\ via Neural Positive-Unlabeled Learning}

\author{\textbf{Alon Jacovi\textsuperscript{\rm 1}\thanks{ \ \ \ Most of the work done during internship at RIKEN.} \ \;\;\; Gang Niu\textsuperscript{\rm 2}\;\;\; Yoav Goldberg\textsuperscript{\rm 1,3}\;\;\; Masashi Sugiyama\textsuperscript{\rm 2,4}} \\ 
\textsuperscript{\rm 1}Bar Ilan University, Israel\\
\textsuperscript{\rm 2}RIKEN, Japan\\
\textsuperscript{\rm 3}Allen Institute for Artificial Intelligence, Israel\\
\textsuperscript{\rm 4}University of Tokyo, Japan\\ 
\texttt{\{alonjacovi,yoav.goldberg\}@gmail.com}, \\ \texttt{gang.niu@riken.jp},\;\; \texttt{sugi@k.u-tokyo.ac.jp} 
}

\date{}

\begin{document}
\maketitle
\begin{abstract}
    We consider the situation in which a user has collected a small set of documents on a cohesive topic, and they want to retrieve additional documents on this topic from a large collection. Information Retrieval (IR) solutions treat the document set as a query, and look for similar documents in the collection. We propose to extend the IR approach by treating the problem as an instance of positive-unlabeled (PU) learning---i.e., learning binary classifiers from only positive (the query documents) and unlabeled (the results of the IR engine) data.
    Utilizing PU learning for text with big neural networks is a largely unexplored field. We discuss various challenges in applying PU learning to the setting, showing that the standard implementations of state-of-the-art PU solutions fail. 
    We propose solutions for each of the challenges and empirically validate them with ablation tests. We demonstrate the effectiveness of the new method using a series of experiments of retrieving PubMed abstracts adhering to fine-grained topics, showing improvements over the common IR solution and other baselines.
\end{abstract}

\section{Introduction}

We are interested in the task of \emph{focused document set expansion}, in which a user has identified a set of documents on a focused and cohesive topic, and they wish to find more documents about the same topic in a large collection. This problem is also known as a ``More Like This'' (MLT) query in web retrieval. A common way of modeling this problem is to consider the set of documents as a long \emph{query}, with which Information Retrieval (IR) techniques can rank documents. IR literature on document similarity and ranking is vast \cite[inter alia]{Faloutsos:1995:SIR:222929,Mitra:2000:IRD:593956.593986}---beyond the scope of this work, and largely orthogonal to it, as will be explained later. 

Current methods in document set expansion for very large collections are based on word-frequency or bag-of-words document similarity metrics such as Term Frequency-Inverse Document Frequency (TF-IDF) and Okapi BM25 and its variants \cite{DBLP:journals/ftir/RobertsonZ09,bm25variant}, considered strong due to their robustness to extreme class imbalance, corpus variance and variable length inputs, as well as their scalability and efficiency \cite{DBLP:journals/corr/MitraC17}. However, the performance of such solutions is limited, as the models cannot capture local or global relationships between words.

We examine methods to improve document set expansion by leveraging non-linear (neural) models under the setting of imbalanced binary text classification. To this end, we look to positive-unlabeled (PU) learning \cite{DBLP:conf/icml/PlessisNS15}: a binary classification setting where a classifier is trained based on only positive and unlabeled data. In the standard document expansion setting, we indeed only possess positive (the document set) and unlabeled (the very large collection) data.

PU learning has originally been employed for text classification by 
\newcite{DBLP:conf/icml/LiuLYL02}; \newcite{DBLP:conf/ecml/LiL05}; \newcite{DBLP:conf/ijcai/LiL03} by using techniques such as EM and SVM. Since then, the setting has been well studied theoretically \cite{DBLP:conf/kdd/ElkanN08,DBLP:conf/icml/PlessisNS15,DBLP:conf/nips/NiuPSMS16}, and recently objective functions have been developed to facilitate training of flexible neural networks from PU data \cite{DBLP:conf/nips/KiryoNPS17}.  We discuss the PU learning setting in more detail in Section \ref{sec:background}, and relevant work on PU learning for text in Section \ref{sec:related-work}.

We are, however, not interested in replacing traditional (term-frequency-based) IR solutions, but rather improve upon their results by further classifying the outputs of those models. There are two reasons for this approach:
    (1) Traditional IR engines are based on word frequencies, and as a result, cannot capture features based on word order;
    (2) Classification by the use of neural networks does not scale well to ``extreme'' imbalance\footnote{In practice, an IR task may involve positive documents in the order of hundreds or thousands, and negative documents in the order of dozens of millions. Literature dealing with imbalanced classification traditionally discuss typical ratios of 1:50 and 1:100 \cite{DBLP:journals/corr/abs-1806-00194,DBLP:journals/corr/abs-1804-10851}. To our knowledge, the setting of extreme imbalance has not been discussed in literature.}.

Following these observations, we see traditional IR engines and neural models as complementary to each other. Our proposed solution is a two-step process: 
    First, a BM25-based, MLT IR engine retrieves relevant candidates;
    Then, a non-linear PU learning model is trained based on the subset of candidates.
In this way, each method relieves the weakness of the other.

As already discussed above, PU learning has recently become viable for deep neural network models. As a result, we are able to leverage it to train models that are able to capture higher order features between words. However, PU learning literature focused on theoretical analysis and experiments on small models and simple---\textbf{notably, class-balanced}---benchmarks such as MNIST, CIFAR10 and 20News \cite{kato2018learning,DBLP:journals/corr/abs-1810-00846,DBLP:journals/corr/abs-1901-10155}. PU learning has not been extensively tested for imbalanced datasets. Scaling PU solutions to high-dimensional, ambiguous and complex data is a significant challenge. One reason for this is that PU data is, by definition, difficult or sometimes impossible to be fully labeled for exhaustive, large-scale evaluation.

For the purpose of document set expansion, and in particular for fine-grained topics, gathering fully-labeled data for an accurate benchmark is also a challenge. For this reason, we propose to simulate the scenario synthetically but realistically by using the PubMed collection of bio-medical academic papers.
PubMed entries are manually assigned multiple terms from Medical Subject Headings (MeSH), a large ontology of medical terms and topics.  We can treat a set of MeSH terms as defining a fine-grained topic, and use the MeSH labels for deriving fully-labeled tasks (see examples of MeSH topic conjunctions in Table \ref{tab:pu-vs-baselines}).
This results in an evaluation setup which is \textbf{extensive}, allowing for a large variety of different datasets based on different bio-medical topics; \textbf{flexible}, with the ability to simulate different biases in the data gathering to account for many possible practical settings; and \textbf{accurate}, with a fully labeled test set. 

The contributions of this work are thus:\footnote{Our code and data are available online at 

\url{https://github.com/alonjacovi/document-set-expansion-pu}.}

\begin{enumerate}
    \item We propose a procedure for generating DSE tasks based on PubMed by using conjunctions of MeSH terms for labels. This serves as a new large-scale benchmark for evaluating (1) PU learning solutions, and (2) DSE solutions, both of which currently suffer from lack of difficult and large-scale evaluation. 
    \item We expand the PubMed DSE task formulation with a variant that includes biased or unbiased negative data.
    \item We apply state-of-the-art PU solutions, previously only evaluated on simple benchmarks for small neural networks, to the PubMed DSE tasks, and report that they perform poorly due to various challenges: no knowledge of class prior, batch size restrictions, extremely imbalanced data (small class prior), and very limited labeled data.
    \item We propose methods to deal with the above challenges, and empirically evaluate the new PU solution (which incorporates our solution) on the PubMed DSE tasks, noting a significant improvement over currently available methods.
\end{enumerate}

\begin{table*}[t]
\centering
\ra{1.25}
\resizebox{0.99\textwidth}{!}{ 
    \begin{tabular}{@{}clcccccc|c@{}}\toprule
        $|LP|$ & Topic & BM25+nnPU & BM25 & Rand+nnPU & BM25+COPK & Naive & All + & Upperbound \\ \midrule
        \multirow{4}{*}{20} & Animals + Brain + Rats & 48.97 & 32.25 $\pm$ 11.6 & 40.21 & 30.47 & 1.49 & 44.6 & 68.17 \\
        & Adult + Middle Aged + HIV Infections & 42.38 & 26.75 $\pm$ 7.22 & 40.22 & 33.59 & 6.88 & 30.98 & 55.61 \\ 
        & Renal Dialysis + Chronic Kidney Failure + Middle Aged & 49.16 & 41.23 $\pm$ 8.95 & 46.58 & 25.4 & 0.00 & 28.40 & 58.18 \\ 
        \cline{2-9}
         & Average of 10$^{\dagger}$ topics & \textbf{33.26} & 26.69 $\pm$ 7.18 & 30.9 & 25.47 & 2.16 & 26.46 & 50.46 \\ \midrule
        \multirow{5}{*}{50} & Animals + Brain + Rats & 60.56 & 32.8 $\pm$ 10.9 & 45.13 & 30.47 & 5.41 & 45.86 & 70.23 \\
        & Adult + Middle Aged + HIV Infections  & 42.77 & 31.85 $\pm$ 10.7 & 50.52 & 33.59 & 12.28 & 40.53 & 58.10 \\ 
        & Renal Dialysis + Chronic Kidney Failure + Middle Aged & 50.09 & 35.78 $\pm$ 9.13 & 45.37 & 25.43 & 0.00 & 31.81 & 57.58 \\ 
        \cline{2-9}
         & Average of 10$^{\dagger}$ topics & \textbf{37.36} & 29.07 $\pm$ 7.75 & 37.01 & 26.51  & 3.01 & 30.41 & 51.09 \\
         & Average of 15$^{\ddagger}$ topics & \textbf{33.82} & 27.55 $\pm$ 6.20 & 31.08 & 25.93 & 2.12 & 29.02 & 47.41 \\
         \bottomrule
    \end{tabular}
}
\caption{Experiment F1 results against the baselines of average performance across topics, as well as three example topics. See Section \ref{sec:evaluation} for details. $\dagger$ denotes the same collection of topics. The average of 15 topics $\ddagger$ includes $\dagger$. The nnPU experiments include BER optimization and proportional batching, but \emph{without} pre-trained embeddings. All experiments use a $|U|$ size of 20,000.}
\label{tab:pu-vs-baselines}
\end{table*}

\section{Background: Positive-Unlabeled Learning} \label{sec:background}

PU learning refers to learning a binary classifier from positive and unlabeled data. In this section we briefly describe notation and relevant literature.

\paragraph{Notation.} We refer to the \textit{positive set} as \textit{P}, the \textit{labeled positive set} as \textit{LP}, the \textit{unlabeled set} as \textit{U}, and the \textit{negative set} as \textit{N}. Empirical approximations of expectations and priors are denoted   $\ \widehat{\cdot}$ .

\subsection{Setting}


Let $x \in \mathbb{R}^d$ and $y \in \{+1, -1\}$ be random variables jointly distributed 
by $p(x,y)$ where $p^+(x) := p(x \mid y = +1)$ and $p^-(x) := p(x \mid y = -1)$ are the 
class marginals (i.e., the positive and negative class-conditional densities), and let 
$g: \mathbb{R}^d \rightarrow \mathbb{R}$ and $\ell: \mathbb{R} \times \{\pm1\} 
\rightarrow \mathbb{R}_+$ be an arbitrary binary decision function and a loss function 
of $(g(x),y)$ respectively. For the purpose of this work, we will use the common sigmoid 
loss, $\ell_{\text{sig}}(t, y) = \frac{1}{1 + \text{exp}(ty)}$, as we have 
observed the best empirical performance with this loss. We denote $\pi^+ := p(y=+1)$ and 
$\pi^- := p(y=-1)$ as the class prior probabilities, such that $\pi^+ + \pi^- = 1$. The methods described in this section all assume the proportion $\pi^+$ to be known.

Binary classification aims to minimize the risk: 
\vspace{-0.1cm}
$$R(g) := \mathbb{E}_{(x,y) \sim 
p(x,y)}[\ell (g(x),y)].$$ 

In supervised (positive and negative: \textit{PN}) learning, both positive $P := \{x^P_i\}_{n^+} \sim p^+(x)$ and negative $N := \{x^N_i\}_{n^-} \sim p^-(x)$ samples are available. The supervised classification risk can be expressed as the partial class-specific risks:
\vspace{-0.2cm}
\begin{multline} \label{step1}
    R(g) =
    \pi^+ \mathbb{E}_{x \sim p^+(x)}[\ell (g(x),+1)] \\
    + \pi^- \mathbb{E}_{x \sim p^-(x)}[\ell (g(x),-1)]. 
\end{multline}

Notice that under the zero-one loss ($\ell_{01}$), the risk $R(g)$ refers to $\pi^+\frac{FN}{FN+TP} + \pi^-\frac{FP}{TN+FP}$. When training, we use $\ell_{\text{sig}}$ which can be regarded as a soft approximation of this formulation for back-propagation.
In practice, the expectations are expressed as the average of losses and optimized in batched gradient-descent or similar methods.

\subsection{Unbiased PU Learning (uPU)}

We utilize the case-control variant of PU learning\footnote{In case-control PU learning, the positive and unlabeled data are collected separately. There are other variants which assume different distributions on the data.} \cite{Ward2009PresenceonlyDA}. Formally, unlabeled data $U := \{x^U_i\}_{n^u} \sim p(x)$ is available instead of $N$, in addition to $P = \{x^P_i\}_{n^+} \sim p^+(x)$ as before.

In order to train a binary classifier from PU data, we could naively train a classifier to separate positive from unlabeled samples. This approach will result, of course, in a sub-optimal biased solution since the unlabeled dataset contains both positive and negative data.  \newcite{DBLP:conf/icml/PlessisNS15} proposed the following unbiased risk estimator to train a binary classifier from PU data.

Since 
\vspace{-0.2cm}
\begin{multline*} \pi^- \mathbb{E}_{x \sim p^-(x)}[f(x)] = \\ \mathbb{E}_{x \sim p(x)}[f(x)] - \pi^+ \mathbb{E}_{x \sim p^+(x)}[f(x)], \end{multline*}
we can substitute the negative-class expectation in Equation (\ref{step1}):
\vspace{-.1cm}
\begin{multline} \label{eq:upu-risk-background}
    R_{PU}(g) := \mathbb{E}_{(x,y) \sim p(x,y)}[\ell (g(x),y)] = \\
    \pi^+ \mathbb{E}_{x \sim p^+(x)}[\ell (g(x),+1)] \\
    + \mathbb{E}_{x \sim p(x)}[\ell (g(x),-1)] \\
    - \pi^+ \mathbb{E}_{x \sim p^+(x)}[\ell (g(x),-1)].
\end{multline}
By empirically approximating this risk as an average of losses over our available dataset, we arrive at an unbiased risk estimator that can be trained on PU data, referred to as the \textit{uPU} empirical risk.


\paragraph{Non-negative PU (nnPU).} If the loss $\ell$ is always positive, so should be the risk. However, \newcite{DBLP:conf/nips/KiryoNPS17} noted that by using stochastic batched optimization, and specifically via very flexible models (such as neural networks), the negative portion of the uPU loss can eventually cause the loss to go negative during training. To mitigate this overfitting phenomenon, they proposed to encourage the loss to stay positive by using gradient-ascent on the negative portion (which replaces the negative-class risk of the classification risk) when it becomes negative. This method is referred to as \textit{nnPU}.

\section{The PubMed Set Expansion Task} \label{sec:pubmed-task}

In this section we discuss the method of generating an extensive benchmark for evaluating solutions of MLT document set expansion.

We are inspired by the following scenario: A user has a set of documents which all pertain to a latent topic, and is interested in retrieving more documents about that topic from a large collection. While traditional term-frequency-based IR solutions scale well to extremely large collections of documents, they are imprecise, and contain a significant amount of noise. Therefore, an additional step based on PU learning can be utilized to classify the output of the IR model, and improve the results.

We are interested in gathering a task for evaluation of the second step. In other words, given an existing black-box IR solution, we would like to use it to produce a dataset for training and evaluation of models which should improve upon the black-box IR solution's performance.

Due to the varied nature of the setting, it is impractical to acquire full supervision for a large number of topics. Therefore, we propose to generate synthetic tasks inspired by the real use-case application.

\subsection{Task Generation Method}

We generate the document-set expansion tasks by
leveraging the expansive PubMed Database: A collection of 29 million bio-medical academic papers. Each document is labeled with MeSH tags, denoting the subject of the document. A conjunction of MeSH terms defines a fine-grained topic, which we use to simulate a user's information intent (example conjunctions in Table \ref{tab:pu-vs-baselines}).

The method of generating one task is then:
\begin{enumerate}
    \item \textit{Input}: $T \leftarrow$ set of MeSH terms (the retrieval topic); $n^+ \leftarrow$ number of labeled positive data; $\text{IR}, \theta_T \leftarrow$ a black-box MLT IR engine, along with query parameters.
    \item $\text{LP} \leftarrow n^+$ randomly selected
    papers that are labeled with $T$.
    \item $U \leftarrow \text{IR}(\text{LP}; \theta_T)$.
\end{enumerate}
For the tasks generated and utilized in this paper, we have chosen MeSH sets manually, and $n^+ \in \{20, 50\}$ (for the training set). For the MLT IR engine we have used the Elasticsearch\footnote{https://www.elastic.co/} implementation of BM25. The top-$\{10000, 20000\}$ scoring documents are retrieved. We make use of the abstracts of the PubMed papers only. See Appendix A for exact details of our method, as well as a comparison to an alternative method for generating \textit{censoring PU} (explained in the appendix) tasks.\footnote{The code for generating the tasks, and the data of our generated tasks are available online at the aforementioned repository. The uploaded dataset contains the paper abstracts. The PubMed identifiers are also available in cases where additional information about each paper, such as the full text, can be retrieved from PubMed if desired.}.

We note that although in essence document set expansion involves using $U$ for both training and evaluation (\textit{transductive case}), we are interested in the case where the PU model is able to generalize to unseen data (\textit{inductive case}). As a result, we split the dataset $[LP; U]$ into training, validation, and test sets, where we use the validation set for hyper-parameter tuning and early-stopping, and evaluate on the test set using the true labels. In other words, we assume a separate (from training) small PU set is available for validation. In our experiments, the size of the validation set is half of the size of the training set. In a deployment setting, the PU model can be used to label the training $U$ data.

\section{Experiment Details}

The rest of this work will reference experiment results. Unless otherwise noted, our base architecture is a single-layer CNN \cite{DBLP:conf/emnlp/Kim14}. The choice of CNN, over other recurrent-based or attention-based models, is due to this architecture achieving the best performance in our experiments.
Test-set performance is reported as an average over multiple MeSH topics (as many as our resources allowed). Except for the experiments that use pretrained models, the inputs are tokenized by words, and word embeddings are randomly initialized and trained with the model. More details are available in Appendix B. We stress that our intent in this work is not to report the very best scores possible, but rather to perform controlled experiments to test hypotheses. To this end, many orthogonally beneficial ``tricks'' in NLP literature were not utilized. Additionally, nnPU-trained models generally required more diligent hyperparameter tuning due to an additional two hyperparameters.

\section{PU Learning for Document Set Expansion} \label{sec:pu-for-retrieval}

In PU classification literature, traditionally small (and in many cases, linear) models have been used on relatively simple tasks, such as CIFAR-10 and 20News. However, performance of existing methods does not scale well to very high-dimensional inputs and state-of-the-art neural models for text classification; applying the PU learning methods described in Section \ref{sec:background} to a more practical setting results in several critical challenges that must be overcome---for example,  PU learning methods often assume a known class prior, yet estimation of the class prior, particularly for text, is hard and inaccurate. In this section we discuss various challenges we have encountered in applying PU learning to the PubMed Set Expansion task, along with proposed, empirically validated solutions.

\begin{table}[t]
    \centering
    \ra{1.25}
    \resizebox{0.55\linewidth}{!}{%
    \begin{tabular}{@{}cccc@{}}\toprule
         $|LP|$ & Prior & Accuracy & F1 \\ \midrule
         20 & $\pi^+$ & 84.27 & 0.0 \\
         20 & 0.5 & 62.09 & 33.26 \\
         50 & $\pi^+$ & 81.71 & 0.0 \\
         50 & 0.5 & 59.92 & 37.36\\
         \bottomrule
    \end{tabular}}
    \caption{Experiments for the PU model, trained with the nnPU loss with either the true class prior (optimizing for accuracy surrogate) or a prior of 0.5 (optimizing for BER surrogate). Reported average across 10$^\dagger$ topics.}
    \label{tab:unknown-prior}
\end{table}

\subsection{Class Imbalance and Unknown Prior (BER Optimization)} \label{subsec:unknown-prior}

Due to the class imbalance (very small class prior), the classification risk encourages the model to be biased towards negative-class prediction (by prioritizing accuracy) in lieu of a model that achieves worse accuracy but better F1. Thus, optimizing for a metric that is similar to F1 or AUC is preferable.

Under a known class prior $\pi^+$ assumption, \newcite{DBLP:journals/ml/SakaiNS18} derived a PU risk estimator for optimizing AUC directly. However, $\pi^+$ cannot be assumed to be known in practice. Furthermore, the high dimensionality and lack of cluster assumption in the input of our task makes estimation difficult and noisy \cite{pmlr-v37-menon15,DBLP:conf/icml/RamaswamyST16,DBLP:conf/nips/JainWR16,DBLP:journals/ml/PlessisNS17}.

Following this line of thought, we propose a simple solution to both problems: by assuming a prior of $\widehat{\pi}^+ = 0.5$ in the uPU loss regardless of the value of the true prior, we are able to optimize a surrogate loss for the Balanced Error (BER) metric\footnote{Given a decision function $g$: $$BER(g) = \frac{1}{2}(\frac{FP}{TN+FP} + \frac{FN}{FN+TP})$$ $$R(g;\ell_{01}) = \pi^-\frac{FP}{TN+FP} + \pi^+\frac{FN}{FN+TP}$$}
\cite{DBLP:conf/icpr/BrodersenOSB10}. Effectively, the uPU loss we are optimizing is: 
\vspace{-0.2cm}
\begin{multline} \label{upu-risk}
    R_{PU}(g) = \\
    \frac{1}{2} \mathbb{E}_{x \sim p^+(x)}[\ell (g(x),+1) - \ell (g(x),-1)] \\
    + \mathbb{E}_{x \sim p(x)}[\ell (g(x),-1)].
\end{multline}

When using the zero-one loss ($\ell_{01}$), the binary classification risk is equivalent to BER, while BER minimization is equivalent to AUC maximization: $AUC = \frac{3}{2} - 2BER$ \cite{pmlr-v37-menon15}. Since back-propagation requires a surrogate loss in place of $\ell_{01}$, such as $\ell_{\text{sig}}$, the BER and AUC metrics are not inversely equivalent; However, we've found BER optimization to perform well in practice.

\paragraph{Results.} Table \ref{tab:unknown-prior} shows a performance comparison in which the models trained using a prior of $0.5$ achieved stronger F1 performance despite weaker accuracy.

\subsection{Small Batch Size (Proportional Batching)} \label{sec:proportional-batching}

The large memory requirements of state of the art neural models such as Transformer \cite{DBLP:conf/nips/VaswaniSPUJGKP17} and BERT \cite{DBLP:journals/corr/abs-1810-04805}, as discussed in the next subsection, coupled with the need to
run on GPU, restrict the batch sizes that can be used. 

This presents a challenge: When the loss function is composed of losses for multiple classes,
when using stochastic batched optimization, each batch should contain a proportionate amount of data of each class relative to the entire dataset. When the classes are greatly imbalanced, this imposes a lower-bound on the batch size when the batch contains one positive example or more. For example, for a dataset which contains 50 positive and 10,000 unlabeled samples, each batch which contains a positive sample must have 200 unlabeled samples. In practice, we were limited to the vicinity of 20 samples per batch when training large Transformer models.

Using a smaller batch-size than the lower-bound (in the case of the example, 20 as opposed to 201) implies that the vast majority of batches will not have labeled positive samples. This result damages performance in multiple ways. First, the model may overfit to the unlabeled data: Since unlabeled examples are treated as discounted negative examples by the uPU loss, the model will be encouraged to predict the negative class due to an abundance of batches that contain only the ``negative'' (in truth unlabeled) class. Additionally, early-stopping may be compromised due to the significantly smaller loss in batches that contain only unlabeled data. 

To solve these problems, we propose to increase the sampling frequency of the positive class \emph{inversely} to its frequency in the dataset. In practice, this solution simply enforces each batch to have a \emph{rounded-up} proportion of its samples for each class. In the example above, every batch with 20 samples will have 1 positive and 19 unlabeled samples. As we ``run out'' of positive samples before unlabeled samples, we define an epoch as the a single loop through the positive set. 

The implication of increasing the sampling frequency is essentially that the positive component of the uPU loss receives a stronger weight. In our running example, the sampling frequency was increased $\times10$. For a sampling frequency increase by an order of $\alpha$, the uPU loss becomes:
\vspace{-0.2cm}
\begin{multline} 
    \grave{R}_{PU}(g) = \\
    \alpha \pi^+ \mathbb{E}_{x \sim p^+(x)}[\ell (g(x),+1)-\ell (g(x),-1)] \\
    + \mathbb{E}_{x \sim p(x)}[\ell (g(x),-1)].
\end{multline}
This, intuitively, counter-acts the overfitting problem caused the abundance of stochastic update steps of entirely unlabeled-class batches. The issue of unstable validation uPU loss is solved as well, since every batch must contain both positive and unlabeled samples, by a ratio that is consistent between the training and validation sets (and thus the validation uPU loss remains a reliable validation metric).

The issue of overfitting in this case is derived from a more general problem: Overfitting to the ``bigger'' class in stochastic optimization of extremely imbalanced data, when the loss can be decomposed into multiple components for each of the classes (as is the case for cross-entropy loss, as well). For this reason, our solution improves ordinary imbalanced classification under batch size restrictions, as well.

\begin{table}[t]
    \centering
    \ra{1.25}
    \resizebox{0.99\linewidth}{!}{%
    \begin{tabular}{@{}ccccc@{}}\toprule
        Setting & Class Ratio & Batch Size & Proportional Batching & F1  \\ \midrule
        \multirow{3}{*}{PN} & \multirow{3}{*}{(P:N) 15:85} & 512 & & 32.55 \\
        & & 16 & & 5.55 \\
        & & 16 & \checkmark & 41.61 \\ \midrule
        \multirow{3}{*}{PU} & \multirow{3}{*}{(LP:U) 2:100} & 512 & & 22.77 \\
        & & 16 & & 0.0 \\
        & & 16 & \checkmark & 22.35 \\
        \bottomrule
    \end{tabular}}
    \caption{Evaluation for the sampling frequency increase method for mitigating overfitting to the bigger class in imbalanced classification with small batch size. Results show that proportional batching dramatically improves results under batch size constraints for both ordinary supervision (PN) and PU settings.}
    \label{tab:batch-size}
\end{table}

\paragraph{Results.} Table \ref{tab:batch-size} shows the effect of the increased sampling frequency method in ordinary imbalanced binary classification, as well as in nnPU training. In the small batch size experiments, the method causes an increase in recall, showing that the model is less inclined towards the ``bigger'' (in our case, the negative) class. \emph{The results apply in both the PN and PU settings, showing that proportional batching can be beneficial to any imbalanced classification task under batch size restrictions.}


\subsection{Limited Data} \label{sec:pretraining}

A defining challenge of document set expansion tasks, when observed through the lens of imbalanced classification, is the very small class prior and small amount of labeled positive data.
Although BER optimization mitigates the issue of the class imbalance, the issue of very little labeled data remains. To this end, we investigate pretraining as a solution. 

We utilize SciBERT \cite{DBLP:journals/corr/abs-1903-10676} for pretrained contextual embeddings in the PubMed domain. For PubMed abstracts that go above the 512 word-piece limit of SciBERT, we utilize a sliding-window approach that averages all embeddings for a word-piece that appeared in multiple windows.

\paragraph{Results.} Utilizing SciBERT embeddings has yielded an increase of F1 performance from 25.75 to 29.96 as an average of five topics. 


\section{Effectiveness of PU Learning} \label{sec:evaluation}

In this section we evaluate the viability of our proposed solution. All experiments in this section use \emph{BER optimization} and \emph{proportional batching} (Section 
\ref{sec:pu-for-retrieval}), but no pre-trained embeddings. We refer to our proposed method as \textbf{BM25+nnPU} where the IR solution BM25 selects the unlabeled dataset for the PU solution, a CNN model with the nnPU loss. 

As an anchor for comparison, we use the following \textit{reference}: \textbf{Upper-bound}: An identical model, trained on the same training data with full supervision using the true labels. This reference can be regarded as the upper-bound performance in the ideal case.

We directly compare against the current commonly-used and best-performing solution as \textbf{IR (BM25)}\footnote{We note that the comparison here should be made to the \emph{specific IR engine} which resulted in the dataset of the PU model, as the PU model benefits greatly from better performance in the IR engine.}: The top-$k$ documents of the IR engine's output, for $k \in \{i\}_{i=|LP|}^{5000}$, are selected as positive documents, while the rest are treated as negative. F1 mean and standard deviation are reported across $k$. \textbf{This strong baseline serves as a reference to the state-of-the-art.}

We additionally compare against standard DSE baselines \textbf{All + (all positive)}: Classifying all samples as the positive class; and \textbf{Naive}: Supervised learning between the labeled positive set (as P) and the unlabeled set (as N).

Finally, we compare against two additional baselines with the aim of validating the beneficial synergy between the IR step and the PU step. In the \textbf{Rand+PU} baseline, we replace the IR step with a random selection of U data. In the \textbf{BM25+COPK} baseline, we replace the PU step with a \textit{Constrained K-means Clustering} \cite{DBLP:conf/icml/WagstaffCRS01} solution, where we perform $k$-means clustering, $k=2$, under the constraint that all $LP$ examples must be in the same cluster. To represent examples in embedding space for $k$-means, we encode the text with SciBERT. Additional details of constrained clustering as a replacement to PU learning are discussed in Appendix \ref{appendix:cop-kmeans}.

The IR baseline is the main alternative to our approach. The all-positive and naive baselines are very simplistic ``lower-bound'' models to be compared against, while the other two-step baselines evaluate the IR or PU steps separately, providing further justification to using the IR and PU solutions together.


Experiments in Table \ref{tab:pu-vs-baselines} show a significant increase in F1 performance as an average across many topics, against all baselines. 

\begin{figure}[t]
\centering
\includegraphics[width=0.4\textwidth]{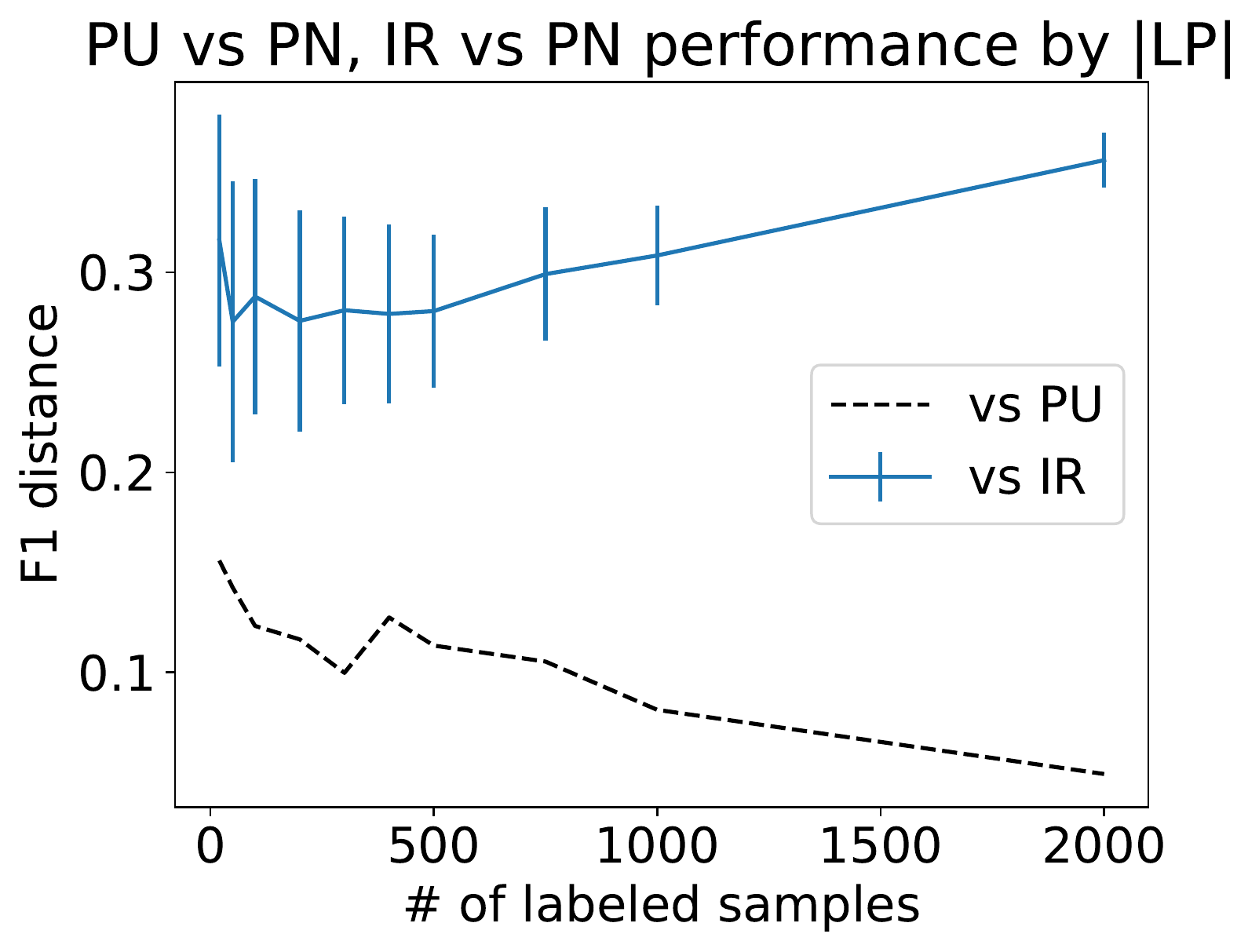}
\caption{The F1 \textit{absolute difference}, normalized by the sum of the two F1 scores, between the upper-bound and nnPU as a function of the amount of labeled positive samples, as well as between the IR top-$k$ baseline (mean and standard deviation) and the upper-bound. Numbers are the average of five topics.
}
\label{fig:pn-vs-pu-diff}
\end{figure}

An interesting experiment in Figure \ref{fig:pn-vs-pu-diff} shows the performance of the IR and PU models normalized by the performance of the upper-bound, as a function of the amount of labeled data. The reported values are the distance of F1 scores between each respective model with the upper-bound, normalized by the sum of scores. The figure shows that as more labeled data is added, the PU model (in truth IR+PU) increases in performance at a rate that is higher than the performance increase of the upper-bound. In comparison, the IR model improvement stays relatively constant beyond 300 labeled samples, while the upper-bound continues to increase, causing the disparity between them to grow. This experiment shows that the \textbf{IR+PU system scales well with increase in \textit{LP} data, increasing performance at a stronger pace than the fully-supervised reference}, while the IR solution scales poorly.

\section{Using Negative Data}

The document set expansion scenario may allow for cases where a limited amount of negative data can be collected. For example, the user may possess some number of relevant negative documents which were acquired alongside the positive documents, prior to training; alternatively, the user may label some documents from the model's output as they appear. Therefore, it is of interest to augment the task with biased labeled negative data---i.e., negative documents which were not sampled from the true negative distribution, but were selected with some bias, such as their length, popularity (for example, the number of citations), or their placement within the IR engine's rankings. 
We consider a bias from document character length, randomly sampling abstracts that are below a certain amount of characters. Alternative bias methods are discussed in Appendix A.

\paragraph{PNU Learning.} When it is possible to obtain negative data in limited capacity, it can be incorporated in training. When the negative data is sampled simply from $p^-(x)$, i.e., it is \textit{unbiased} negative data, it is possible to use \textit{PNU classification} \cite{DBLP:conf/icml/SakaiPNS17}, which is a linear combination of $R(g)$ and $R_{PU}(g)$:
\begin{equation} \label{eq:pnu-loss}
     R_{PNU}(g) := \gamma R(g) + (1-\gamma) R_{PU}(g).
\end{equation}

We note that to our knowledge, PNU learning has not yet been successfully applied to deep models prior to this work. We apply the same solution to the case of biased negative samples. Our PNU experiments include Proportional Batching to overcome the extreme class imbalance.


\begin{table}\centering
\ra{1.25}
\resizebox{0.7\linewidth}{!}{%
\begin{tabular}{@{}lccc@{}}\toprule
Setting & Precision & Recall & F1\\ \midrule
PU & 29.35 & 71.83 & 40.78 \\
PN (unbiased N) & 33.83 & 70.40 & 42.14\\
PN (biased N) & 19.34 & 90.62 & 31.29\\
\bottomrule
\end{tabular}
}
\caption{Experiments for five topics. All experiments used $|LP|=50$, $|N|=50$ for training and  $|LP|=25$, $|N|=25$ for validation (as well as $U$ in the PU setting).}
\label{tab:pnu1}
\end{table}

\begin{table}\centering
\smallskip
\ra{1.25}
\resizebox{0.99\linewidth}{!}{%
\begin{tabular}{@{}llcc@{}}\toprule
Setting & Model & (a) Unbiased N F1 & (b) Biased N F1\\ \midrule
PNU & Ensemble (PN+PU 1+1) & 42.31 & 37.63\\
PNU & Multi-task & 41.50 &  41.48\\
PU & Ensemble (3) & \multicolumn{2}{c}{41.25} \\
\bottomrule
\end{tabular}
}
\caption{Average performance of the same five topics as in Table \ref{tab:pnu1}. All experiments used $|LP|=50$, $|N|=50$ for training and  $|LP|=25$, $|N|=25$ for validation (as well as $U$). Bias selection for N was performed by character length. ``Multi-task'' refers to Equation (\ref{eq:pnu-loss}).}
\label{tab:pnu2}
\end{table}

\paragraph{Results.} Tables \ref{tab:pnu1} and \ref{tab:pnu2} summarize the results of PNU learning for the biased and unbiased settings. We observe that performance improves with unbiased negative samples, but does not improve with negative documents selected with bias to shorter documents. In the unbiased case, a simple ensemble of PN and PU models out-performs PNU learning, and we verify that the ensembling is not the sole cause of the performance increase by noting that the PN and PU ensemble out-performs a 3-model PU ensemble. In the biased case, the performance of the PN model is severely lower than the PU model, and in this case indeed the PNU model slightly out-performs the PN and PU ensemble.


\section{Related Work} \label{sec:related-work}

Linear PU models have been extensively used for text classification \cite{DBLP:conf/aaai/LiuLLY04,DBLP:conf/micai/YuZP05,DBLP:conf/dasfaa/CongLWL04,DBLP:conf/ecml/LiL05} by using EM and SVM algorithms. Particularly, the 20News corpus has been often leveraged to build PU tasks for evaluation of those models \cite{DBLP:conf/icml/LeeL03,DBLP:conf/ijcai/LiLN07}. \newcite{DBLP:conf/acl/LiZLN10} have evaluated EM-based PU models against distributional similarity for entity set expansion.
\newcite{DBLP:conf/emnlp/LiLN10} proposed that PU learning may out-perform PN when only the negative data's distribution significantly differs between training and deployment.

\newcite{DBLP:journals/ml/PlessisNS17}; \newcite{DBLP:journals/corr/abs-1809-05710} describe methods of estimating the class prior from PU data under some distributional assumptions. \newcite{DBLP:journals/corr/abs-1810-00846} introduced \textit{PUbN} as another PU-based loss for learning with biased negatives. PUbN involves two steps, where the marginal probability of a sample to be labeled (positive/negative) is estimated using a neural model, and then used. In our experiments, PUbN has consistently overfit to the majority baseline. We suspect that this is a result from noisy estimation of the labeling probability due to the difficulty of the task.

\section{Conclusion} \label{discussion}

We propose a two-stage solution to document set expansion---the task of retrieving documents from a large collection based on a small set of documents pertaining to a latent fine-grained topic---as a method of improving and expanding upon current IR solutions, by training a PU model on the output of a black-box IR engine. In order to accurately evaluate this method, we synthetically generated tasks by leveraging PubMed MeSH term conjunctions to denote latent topics. Finally, we discuss challenges in applying PU learning to this task, namely an unknown class prior, extremely imbalanced data and batch size restrictions, propose solutions (one of which---``Proportional Batching''---applies in the general scope of PN imbalanced classification, as we empirically validate), and provide empirical evaluation against multiple baselines which showcase the effectiveness of the approach.

\paragraph{Future Work.} Stronger class prior estimation, through additional task assumptions, may facilitate direct AUC optimization. Additionally, methods of increasing precision may be considered (such as data augmentation or adversarial training).

\section*{Acknowledgments}
AJ and YG have received funding from the European Research Council (ERC) under the European Union's Horizon 2020 research and innovation programme, grant agreement No. 802774 (iEXTRACT).
GN and MS were supported by JST AIP Acceleration Research Grant Number
JPMJCR20U3, Japan.

\bibliographystyle{acl_natbib}
\bibliography{eacl2021}

\appendix

\clearpage

\section{PubMed Set Expansion Task Generation} \label{appendix:pubmed-retrieval}

In this section we discuss details of the PubMed Set Expansion task generation process. 

\paragraph{Parameters.} For this work, we have indexed the January 2019 version of PubMed in an Elasticsearch ver-6.5.4 index. We discard all papers in PubMed that do not have MeSH terms or abstracts (of which there are few). The title and abstract of each paper are tokenized using the Elasticsearch English tokenizer, with term vectors. The title receives a 2.0 score boost during retrieval. For retrieval, we use the Elasticsearch ``More Like This'' query with the default implementation of BM25, and a ``minimum should match'' parameter of 20\%, indicating that papers that do not share a 20\% overlap of terms with the query are dropped. This parameter was controlled in the interest of efficiency, as the query is otherwise very slow. 

Table 6 contains statistics about sample topics.

\begin{table}[h]
\centering
\ra{1.2}
\resizebox{0.99\linewidth}{!}{%
\begin{tabular}{l | c | c | c | c} \toprule
    	Topic & $|LP|$ & $|U|$ & Precision & Recall \\
    	\midrule
    	\multirow{2}{*}{Liver + Rats, Inbred Strains + Rats} & 20 & 10,000 & 17.45 & 15.59 \\
    	& 50 & 10,000 & 16.55 & 14.82 \\ 
    	\multirow{2}{*}{Adult + Middle-Aged + HIV Infections} & 20 & 20,000 & 18.33 & 20.06 \\
    	& 50 & 20,000 & 25.42 & 27.85 \\ \bottomrule
\end{tabular}}
\label{tab:dataset-stats}
\caption{Dataset sizes for two example PubMed Set Expansion tasks based on the given topics, each composed of three MeSH terms. The reported sizes are for the training set. Precision denotes the proportion of P samples in U, and recall denotes the proportion of retrieved P samples from all positive documents in PubMed.}
\end{table}

\paragraph{Censoring PU learning.} An alternative, easier, scenario for the Document Set Expansion task involves the case where the LP data was sampled and labeled from the U distribution, termed \textit{censoring PU learning}. To model this case, the task can be generated in the following way: 
\begin{enumerate}
    \item \textit{Input}: $T \leftarrow$ set of MeSH terms (the retrieval topic); $n^+ \leftarrow$ number of labeled positive data; $IR, \theta_T \leftarrow$ a black-box MLT IR engine, along with query parameters.
    \item $P \leftarrow $ All papers that are labeled with $T$.
    \item $N \leftarrow IR(P; \theta_T)$
    \item $LP \leftarrow n^+$ randomly selected papers in P.
    \item $U \leftarrow [P-LP; N]$
\end{enumerate}
Experimentally, the F1 performance of all the models (PU and PN) was greatly increased for this setting, in comparison to the case-control tasks described in the main work. All methods discussed in this work apply to the censoring setting, as it is a special case of case-control.

\begin{figure}[t]
\centering
\includegraphics[width=0.95\linewidth]{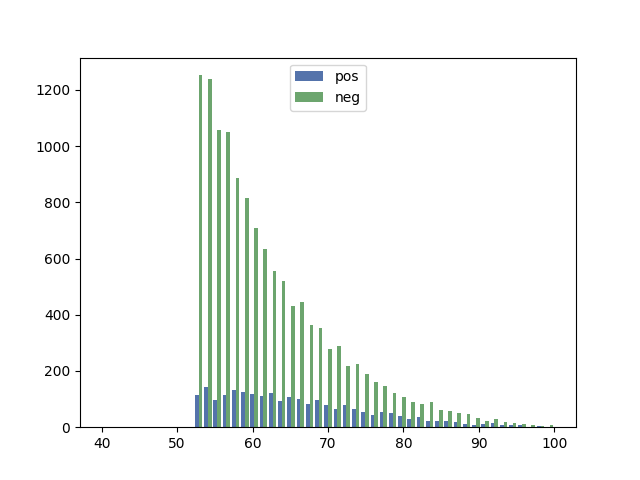}
\caption{Two histograms of $U$ positive and negative documents respectively by their BM25 score. The horizontal axis denotes buckets of BM25 scores, and the vertical axis is the amount of samples in that bucket.}
\label{fig:tfidf-score-histogram}
\end{figure}

\paragraph{Bias.} It is possible to simulate bias in the sampling of documents according to many heuristics and assumptions. For example, it may be assumed that the user is more likely to label documents that are shorter, or documents that are more famous (as indicated by amount of citations in PubMed). Additional possible conditions include the ranking of the IR engine in two possible ways: 1. The user may submit labels after the IR query while viewing the results. In this case, the user is more likely to label documents that are ranked higher; 2. In the case of an IR engine modeled by bag-of-words (such as BM25), documents that rank lower can be assumed to possess less relevant vocabulary overlap with the positive class, such that they may be easier to label at a glance. Figure \ref{fig:tfidf-score-histogram} shows a typical distribution of class according to the rank of BM25 for a sample task of PubMed Set Expansion. 

\section{Experiment Details} \label{appendix:experiments}

The experiments were implemented in PyTorch version 1.0.1.post2, AllenNLP version 0.8.3-unreleased. The neural models used a CNN encoder with max-pooling, with 100 filters for the title and 200 filters for the abstract, split evenly between window sizes of 3 and 5. The choice of CNN (over other recurrent-based or attention-based models) is due to this architecture achieving the best performance in practice. For the SciBERT contextual embeddings, SciBERT-base was used. The learning rate for the model with no pretraining used is 0.001, while the learning rate for the SciBERT model is 0.00005. The nnPU parameters $\beta,\gamma$ were set to 0 and tuned over the validation set loss, respectively. In all cases of nnPU training we used the biggest batch-size possible, which was 1000 for the CNN model with no pretraining, and between 16 to 25 for the SciBERT model. In the case of the SciBERT model, we've ignored training and validation samples longer than 600 words, tokenized by the AllenNLP default implementation of \verb WordTokenizer , to avoid long outliers which greatly limit the batch size. This was not performed on the test set to maintain an unbiased comparison. 

\subsection{Experiment Topics}
The topics were chosen by a policy of related triplets, such that they could conceivably (though loosely) be relevant searches in practice, by sampling and filtering from MeSH triplets that they occur together in PubMed on an order of hundreds, thousands or tens of thousands. The topics were chosen \textit{without} knowledge of any experiment results related to them, such that they were not picked to achieve a particular outcome.
\vspace{0.5cm}

\noindent($\dagger$)
\begin{enumerate}
\item Animals + Brain + Rats.
\item Adult + Middle Aged + HIV Infections.
\item Lymphatic Metastasis + Middle Aged + Neoplasm Staging.
\item Base Sequence + Molecular Sequence Data + Promoter Regions, Genetic.
\item Renal Dialysis + Kidney Failure, Chronic + Middle Aged.
\item Aged + Middle Aged + Laparoscopy.
\item Apoptosis + Cell Line, Tumor + Cell Proliferation.
\item Disease Models, Animal + Rats, Sprague-Dawley + Rats.
\item Liver + Rats, Inbred Strains + Rats.
\item Dose-Response Relationship, Drug + Rats, Sprague-Dawley + Rats.
\end{enumerate}
($\ddagger$)
\begin{enumerate}
\item Female + Infant, Newborn + Pregnancy.
\item Molecular Sequence Data + Phylogeny + Sequence Alignment.
\item Cells, Cultured + Mice, Inbred C57BL + Mice.
\item Dose-Response Relationship, Drug + Rats, Sprague-Dawley + Rats.
\item Brain + Magnetic Resonance Imaging + Middle Aged.
\end{enumerate}

\section{Constrained Clustering for PU Learning} \label{appendix:cop-kmeans}

Unfortunately, we are not aware of many competitive alternative solutions to nnPU that interface with only positive and unlabeled data. One such a solution is \textit{constrained clustering}, or clustering under constraints of prior knowledge on which examples should belong in the same cluster, or which examples should not belong in the same cluster. 

Constrained clustering can be reduced to a PU problem in the following way: Given $LP$ and $U$ data, we perform clustering under constraints that all of the examples in $LP$ must belong in the same cluster. If $N$ data is available, we may constrain all $N$ data to be in the same cluster, as well, and that $LP$ and $N$ examples may not be in the same cluster. If the algorithm allows a parameterization of the number of clusters, such as COP-Kmeans \cite{DBLP:conf/icml/WagstaffCRS01}, we may specify this number to be 2. Otherwise, all clusters that do not contain the $LP$ examples can be selected as clusters of $N$, and the cluster that contains the $LP$ examples shall be selected as $P$.

In this way, we achieve a reduction from the constrained clustering problem to a PU problem, allowing it to serve as a replacement to nnPU. While we are not aware of other work which made this reduction or comparison between constrained clustering and PU learning, in our experiments we note that nnPU has achieved stronger performance and scalability in large data. 

\end{document}